%% file: main.tex
\pdfoutput=1
\documentclass[10pt,twocolumn,letterpaper]{article}

\usepackage[pagenumbers]{cvpr} % To force page numbers, e.g. for an arXiv version

\input{preamble}
\providecommand{\nolinenumbers}{}\providecommand{\linenumbers}{}

\definecolor{cvprblue}{rgb}{0.21,0.49,0.74}
\usepackage[pagebackref,breaklinks,colorlinks,allcolors=cvprblue]{hyperref}
\hypersetup{pdftitle={RoboLiveMem: Is Robot Memory Solved When the World Does Not Wait?},
  pdfauthor={Yunbei Zhang, Yuyou Zhang, Janet Wang, Ming Liu, Yijiang Li, Yikai Wang, Weirui Ye, Ding Zhao, Jihun Hamm}}

\def\confName{CVPR}
\def\confYear{2027}

\title{\bench: Is Robot Memory Solved When the World Does Not Wait?}

\author{
Yunbei Zhang$^{1,\dagger,*}$\quad Yuyou Zhang$^{2,\dagger}$\quad Janet Wang$^{1,\dagger}$\quad Ming Liu$^{3}$\quad Yijiang Li$^{4}$\\
Yikai Wang$^{2}$\quad Weirui Ye$^{5}$\quad Ding Zhao$^{2,*}$\quad Jihun Hamm$^{1,*}$\\[0.5em]
{\normalsize $^{1}$Tulane\quad $^{2}$CMU\quad $^{3}$Amazon\quad $^{4}$UCSD\quad $^{5}$MIT}\\[0.2em]
{\small $^{\dagger}$Equal contribution\quad $^{*}$Co-corresponding authors}\\[0.3em]
{\small \textbf{Project Page \& Code:} \href{https://robolivemem.github.io/}{\textcolor{lmpurpled}{\textbf{https://RoboLiveMem.github.io}}}}
}

\begin{document}
\raggedbottom
\twocolumn[{%
\renewcommand\twocolumn[1][]{#1}%
\maketitle
\vspace{-1.6em}
\begin{center}
  \includegraphics[width=\textwidth]{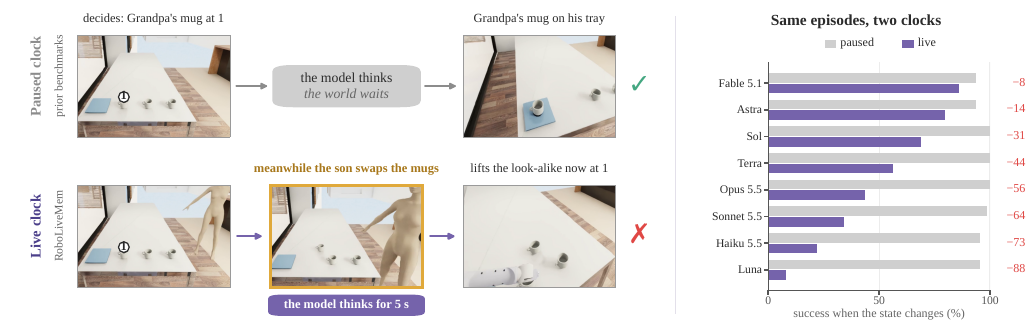}
  \vspace{-1.4em}
  \captionof{figure}{\textbf{Robot memory benchmarks pause the world, and \bench does not.} Left: one recorded episode of
  \emph{Grandpa's mug} (BEHAVIOR-1K, Claude Opus~5.5). Grandpa's mug stands at mark~1 among identical look-alikes. Under
  the paused clock thinking costs no world time and the episode succeeds. Under the live clock the son swaps the mugs
  while the model thinks, the model commits to ``pick~1'' from the frame at 13.0\,s (a) before the swap is over (b,~c),
  and it lifts the look-alike (d), although the swap was in its own camera stream. Right: success of the
  \nfrontier Claude and GPT models on the realistic episodes in which a remembered object changes, with the world paused and
  running, and the drop.}
  \label{fig:teaser}
\end{center}
\vspace{0.6em}
}]

\input{sec/0_abstract}
\input{sec/1_intro}
\input{sec/2_related}
\input{sec/3_livemem}
\input{sec/4_benchmark}
\input{sec/5_protocol}
\input{sec/6_results}
\input{sec/7_conclusion}

\section*{LLM Usage Statement}
We used ChatGPT and Claude to assist with language polishing and improving the readability of this manuscript, and to
give feedback on the manuscript. The vision-language models evaluated in this paper are the objects of study of our
benchmark (\cref{sec:setup}). The authors are fully responsible for all research ideas, experimental results, and claims
presented in this paper.

{
    \small
    \bibliographystyle{ieeenat_fullname}
    \bibliography{main}
}

% Preprint: the appendix follows the references and continues the page numbering of the paper.
\clearpage
\newcounter{lmpage}\setcounter{lmpage}{\value{page}}
\let\lmmaketitlesupp\maketitlesupplementary
\renewcommand{\maketitlesupplementary}{\setcounter{page}{\value{lmpage}}\lmmaketitlesupp}
\input{sec/X_suppl}

\end{document}

%% file: preamble.tex
\usepackage{multirow}
\usepackage{tabularx}
\usepackage{array}
\usepackage{colortbl}
\usepackage{makecell}
\usepackage{pifont}
\usepackage[most]{tcolorbox}
\usepackage{enumitem}

\definecolor{lmpurple}{HTML}{7563AB}
\definecolor{lmpurpled}{HTML}{4B3F8A}
\definecolor{lmlavender}{HTML}{E4DEF5}
\definecolor{lmlav}{HTML}{F5F2FC}
\definecolor{lmamber}{HTML}{A8781C}
\definecolor{lmgreen}{HTML}{44A781}
\definecolor{lmred}{HTML}{E04845}
\definecolor{rowgray}{HTML}{F2F2F2}
\newcommand{\dropc}[1]{\textcolor{lmred}{#1}}  % live drops in tables

\newcommand{\para}[1]{\par\vspace{2pt}\noindent\textbf{#1}\hspace{0.45em}}

\newtcolorbox{finding}[1]{enhanced, breakable=false, width=\linewidth, colback=lmlav, colframe=lmpurple,
  boxrule=0.5pt, arc=1.5pt, left=4pt, right=4pt, top=2.5pt, bottom=2.5pt, before skip=9pt, after skip=9pt,
  fontupper=\small, title={#1}, fonttitle=\small\bfseries, coltitle=white, colbacktitle=lmpurple, toptitle=1pt,
  bottomtitle=1pt, before={\par\nolinenumbers\vspace*{2mm}\noindent}, after={\par\vspace*{1.5mm}\linenumbers}}

\input{tables/counts}

\newcommand{\extramodels}{Gemini~4 Argon and Carbon, Grok~4.7, Amazon Nova~2 Lite, and the open-weight
  Qwen3-VL-235B~\cite{bai2025qwen3vl}, Gemma~3~27B~\cite{gemma2025gemma3}, Kimi K2.5~\cite{kimi2026k25} and
  Ministral~3~14B\xspace}
\newcommand{\bench}{\textsc{RoboLiveMem}\xspace}
\newcommand{\cP}{\textsf{P}\xspace}  % hidden, paused clock
\newcommand{\cL}{\textsf{L}\xspace}  % hidden, live clock
\newcommand{\cV}{\textsf{V}\xspace}  % visible twin, live clock
\newcommand{\lmu}{LM-U\xspace}
\newcommand{\lmw}{LM-W\xspace}
\newcommand{\lmt}{LM-T\xspace}

\newcommand{\tableheader}{\bfseries}
\newcommand{\cmark}{{\color{lmgreen}\ding{51}}}
\newcommand{\xmark}{{\color{lmred}\ding{55}}}
\newcommand{\pmark}{{\color{lmamber}$\circ$}}
\newcolumntype{Y}{>{\raggedright\arraybackslash}X}
\newcolumntype{C}{>{\centering\arraybackslash}X}

\newcommand{\ind}{\mathbf{1}}

\usepackage{microtype}

%% file: tables/counts.tex
\newcommand{\nmodels}{16\xspace}
\newcommand{\nepisodes}{34{,}176\xspace}

\newcommand{\nvendors}{eight\xspace}
\newcommand{\nrealmodels}{16\xspace}
\newcommand{\nrealeps}{5{,}469\xspace}
\newcommand{\nsemimodels}{16\xspace}
\newcommand{\nsemieps}{13{,}774\xspace}
\newcommand{\nctrlmodels}{16\xspace}
\newcommand{\nctrleps}{12{,}053\xspace}
\newcommand{\nrealev}{55--64}

\newcommand{\nfrontier}{8\xspace}
\newcommand{\rPevmin}{94}

\newcommand{\rLevmin}{8}
\newcommand{\rLevmax}{86}
\newcommand{\rPevavg}{97}
\newcommand{\rLevavg}{50}
\newcommand{\rDevmin}{8}
\newcommand{\rDevmax}{88}
\newcommand{\rLnullmin}{92}
\newcommand{\rLnullmax}{100}
\newcommand{\rBest}{Fable~5.1\xspace}
\newcommand{\rWorst}{Luna\xspace}

\newcommand{\rLowPval}{94}
\newcommand{\rhoSemi}{-0.17}
\newcommand{\rhoCtrl}{+0.30}
\newcommand{\rhoReal}{+0.76}
\newcommand{\npairs}{48\xspace}
\newcommand{\nlower}{46\xspace}
\newcommand{\maxgain}{2\xspace}
\newcommand{\cDmin}{9}
\newcommand{\cDmax}{92}
\newcommand{\napi}{eight\xspace}
\newcommand{\aPevmin}{76}
\newcommand{\aPevmax}{98}
\newcommand{\aLevmin}{2}
\newcommand{\aLevmax}{80}
\newcommand{\gArealL}{80}
\newcommand{\gArealLat}{32}

\newcommand{\gCrealL}{73}
\newcommand{\gCrealLat}{56}
\newcommand{\gCctrlL}{6}
\newcommand{\gCctrlLat}{34}

%% file: sec/0_abstract.tex
\begin{abstract}
Robot agents built on frontier vision-language models now solve most robot memory benchmarks. Does this mean they can
remember what they need in a home, where people and machines keep moving things? Existing memory benchmarks cannot
answer this question, because they pause the simulator while the model thinks: nothing can change during a decision, so
whatever the robot remembers is still true when it acts. We introduce \bench, a benchmark that runs the same memory
tasks twice, once with the world paused while the model thinks and once with the world running. It contains 29 task
families in three tiers, from 16 realistic BEHAVIOR-1K scenes of homes, workshops and a restaurant to controlled
ManiSkill tasks. Every task mixes real changes with harmless look-alikes and runs without any change, so that no fixed
strategy succeeds, and most tasks have a twin that keeps the change in view, which separates forgetting from acting
too late.
Testing \nmodels models on \nepisodes episodes, we find three surprises. (1)~Paused results hide the problem: when a
remembered object changes in a realistic scene, the \nfrontier Claude and GPT models succeed in \rPevavg\% of episodes on
average with the world paused, but in only \rLevavg\% with the world running, and their ranking changes: Claude
Opus~5.5, which never fails when paused, succeeds in 44\%. (2)~Speed is not the fix: slower models do no worse, and
models given only 2\,s per decision still fall at least 36 points below their paused success. (3)~Models see
the change but act on camera frames from before it ended. Looking again just before acting closes most of the
realistic gap, yet when identical objects are swapped several times, all nine models we test still lose points to
memory.
\end{abstract}

%% file: sec/1_intro.tex
\section{Introduction}
\label{sec:intro}

A robot that works in a home has to remember what it can no longer see: which of three identical mugs is Grandpa's,
whether the soup has already been salted, where the keys were put. Memory has therefore become a standard test for
robot policies, and several recent benchmarks measure
it~\cite{dai2026robomme,cherepanov2026mikasa,chen2026rmbench,lei2026robomemarena,cherepanov2026mikasavla}. Agents built
on frontier vision-language models now solve most of these tasks, often as well as people
do~\cite{jedlicki2026embodied,wu2026memout,ryu2026memi,zhang2026simplearm}. Robot memory, at least for such agents,
looks nearly solved.

All of these benchmarks share one assumption: the simulator stops while the model thinks. A frontier model needs
seconds for each decision. In a paused world nothing changes during that time, so whatever the robot remembers when it
starts to decide is still true when its command runs. A real home does not wait. In \cref{fig:teaser}, the robot sees
Grandpa's mug at mark~1 and starts to think. Meanwhile the son, who is setting the table, swaps the two mugs. When the
robot acts, its memory is wrong, even though its own camera recorded the swap while it was thinking. With the simulator
paused, the same episode succeeds.

We call the missing ability \emph{live memory}: keeping what the robot remembers correct while people or machines
change it during the robot's own decision. No existing benchmark measures it (\cref{tab:compare}). Memory benchmarks
such as RoboMME~\cite{dai2026robomme}, MIKASA-Robo~\cite{cherepanov2026mikasa}, RMBench~\cite{chen2026rmbench} and
RoboMemArena~\cite{lei2026robomemarena} hide state from the robot, but the world moves only when the policy acts.
Real-time benchmarks such as Realtime-Bench~\cite{tang2026realtimebench}, Gaia2~\cite{froger2026gaia2},
Robo-bench~\cite{vals2026robobench} and a mode of Embodied First Exam~\cite{efe2026} let time run while the model
thinks, but the state that matters is mostly in view when the agent acts, so they test speed rather than memory. When a
model fails in a moving world, there are three possible reasons, and a benchmark has to tell them apart: the model may
be too slow; it may not look at the right moment, since most agents stop watching while they think; or it may lose
track of what changed.

\begin{table}[t]
  \caption{\textbf{\bench and related benchmarks.} Emb.: embodied. Mem.: the task hinges on unobserved state. Run.:
  the world keeps evolving while the agent thinks. Live: other actors change the remembered state after hand-over.
  Ctrl.: decoy and no-change seeds and visible twins. \pmark: partly.}
  \label{tab:compare}
  \centering\footnotesize
  \begin{tabular*}{\columnwidth}{@{\extracolsep{\fill}}lccccc@{}}
    \toprule
    Benchmark & Emb. & Mem. & Run. & Live & Ctrl.\\
    \midrule
    RoboMME~\cite{dai2026robomme}                & \cmark & \cmark & \xmark & \xmark & \xmark\\
    MIKASA-Robo~\cite{cherepanov2026mikasa}      & \cmark & \cmark & \xmark & \xmark & \xmark\\
    RMBench~\cite{chen2026rmbench}               & \cmark & \cmark & \xmark & \xmark & \xmark\\
    RoboMemArena~\cite{lei2026robomemarena}      & \cmark & \cmark & \xmark & \xmark & \xmark\\
    Realtime-Bench~\cite{tang2026realtimebench}  & \xmark & \pmark & \cmark & \xmark & \xmark\\
    Gaia2~\cite{froger2026gaia2}                 & \xmark & \xmark & \cmark & \xmark & \xmark\\
    Robo-bench~\cite{vals2026robobench}          & \cmark & \xmark & \cmark & \xmark & \xmark\\
    Embodied First Exam~\cite{efe2026}           & \cmark & \xmark & \cmark & \pmark & \xmark\\
    \midrule
    \bench (ours)                                & \cmark & \cmark & \cmark & \cmark & \cmark\\
    \bottomrule
  \end{tabular*}
  \vspace{-0.8em}
\end{table}

We build \bench to measure live memory and to separate these three reasons. It runs every task under two clocks: a
paused clock, as in earlier benchmarks, and a live clock that charges the model's thinking time to the world. Eight
checkable conditions keep the live version fair (\cref{sec:livemem}): the change is not announced, the agent does not
trigger it, and real changes are mixed with harmless look-alikes and runs without any change, so that no fixed strategy
wins. Most tasks also have a twin in which the change stays in view, which lets us tell forgetting apart from acting
too late. Its 29 task families come in
three tiers (\cref{fig:tiers}): 16 BEHAVIOR-1K~\cite{li2022behavior1k} scenes in which people, robots and machines make
the changes, a MuJoCo kitchen, and 5 ManiSkill tasks.

We test \nmodels models from \nvendors providers on \nepisodes episodes (\cref{sec:results}) and find the following.
(1)~Paused success does not predict live success. When a remembered object changes in a realistic scene, the
\nfrontier Claude and GPT models succeed in \rPevavg\% of episodes on average with the world paused but in only \rLevavg\%
with the world running. Their paused scores are nearly equal and their live scores are not: Opus, perfect when paused,
succeeds in 44\%, while Fable~5.1, tied for the lowest paused score, succeeds in 86\% (\cref{fig:teaser,fig:scatter}).
The \napi models we run in our own agent also all lose success live. (2)~Speed is not the fix. Slower models do no worse,
and each of nine models given only 2\,s per decision still falls at least 36 points below its paused success
(\cref{fig:speed}). (3)~The models see the change. Asked afterwards about the frames they had, they describe it
correctly; they fail because they act on frames taken before the change was over (\cref{fig:fresh}). Looking again just
before acting recovers most of the realistic gap. (4)~What remains is memory. When identical objects are swapped
several times, each of nine models still does at least 12 points worse than when the swaps stay in view, even when it looks
again before acting (\cref{tab:commit}).

\para{Contributions.} (1)~We define live memory by letting the world run while the model thinks, with eight checkable
conditions for a valid task and a protocol that separates speed, observation and memory. (2)~We build \bench, to our knowledge the first
robot memory benchmark that runs identical tasks with the world paused and running. (3)~We evaluate \nmodels models and
show that paused benchmarks overstate the memory of frontier agents.

%% file: sec/2_related.tex
\section{Related Work}
\label{sec:related}

\begin{figure*}[t]
  \centering
  \includegraphics[width=\textwidth]{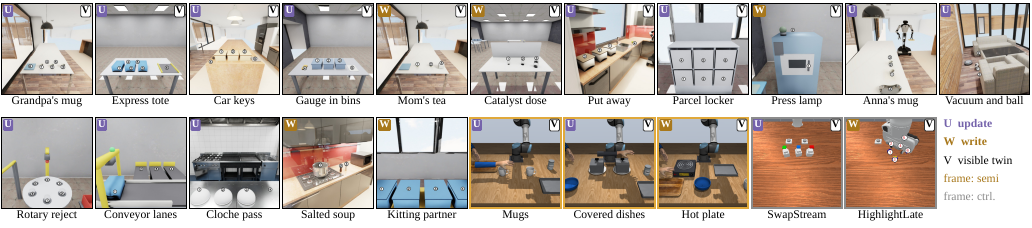}
  \vspace{-1.8em}
  \caption{\textbf{Examples from the three tiers.} The 16 realistic BEHAVIOR-1K scenes, shown as the agent's first view
  with the numbered marks its skills act on, followed by three semi-realistic MuJoCo kitchen families (ochre frame) and two
  controlled ManiSkill tasks in their visible-twin form (gray frame). Badges give the memory class (U update, W write)
  and whether the family has a visible twin (V).}
  \label{fig:tiers}
  \vspace{-0.4em}
\end{figure*}

\para{Robot memory.} Memory benchmarks for manipulation test whether a policy can recall hidden objects, counts,
demonstrated patterns and past events~\cite{dai2026robomme,cherepanov2026mikasa,cherepanov2026mikasavla,chen2026rmbench,lei2026robomemarena},
and RoboMME-Interference~\cite{rathi2026interference} adds distractors to the history. Memory-augmented
policies~\cite{sridhar2026memer,yin2026simplememvla,sun2026ltm} and agentic systems on frontier
models~\cite{jedlicki2026embodied,wu2026memout,ryu2026memi,chen2026astrabank,zhang2026simplearm} report near-human or
above-human scores on these benchmarks. All of them step the simulator in lockstep with the policy, so the remembered
state cannot change while the policy computes. In contrast, \bench keeps their notion of memory and lets other actors
change the remembered state while the agent deliberates.

\para{Embodied agents on foundation models.} Language and vision-language models plan and act through skills, code or
marked images~\cite{ahn2022saycan,huang2022inner,liang2023code,yang2023setofmark,fu2026capx}, and benchmarks such as
EmbodiedBench~\cite{yang2025embodiedbench} and Embodied First Exam~\cite{efe2026} evaluate them across many tasks. These
evaluations are turn-based, except a mode of Embodied First Exam in which time keeps running on largely observable
tasks. We evaluate the same kind of agent, but on tasks whose decisive state is hidden when the agent acts.

\para{Real-time agents and acting while computing.} Realtime-Bench~\cite{tang2026realtimebench}, Real-Time Reasoning
Gym~\cite{wen2025realtime}, Gaia2~\cite{froger2026gaia2} and Robo-bench~\cite{vals2026robobench} show that reasoning
models degrade when the environment does not wait. Real-time chunking~\cite{black2025rtc}, reinforcement learning for
real-time VLAs~\cite{dong2026rtexpo}, PonderPounce~\cite{choi2026ponderpounce}, TacitVLA~\cite{lu2026tacitvla} and
VoLo~\cite{chen2026volo} change the policy so that it acts during inference. These works target latency, the right
answer arriving too late. Different from them, we isolate a robot that acts in time on a memory that became false during
its own deliberation, and we report every harness setting, since harness choices change agent
results~\cite{zhang2026harness}.

%% file: sec/3_livemem.tex
\section{Live Memory: Formulation and Protocol}
\label{sec:livemem}

\begin{table}[t]
  \caption{\textbf{What makes an instance a test of live memory.} Every family is checked against all eight conditions
  before any model runs. Formal statements in \cref{app:checks}.}
  \label{tab:conditions}
  \centering\footnotesize
  \begin{tabular*}{\columnwidth}{@{\extracolsep{\fill}}l l@{}}
    \toprule
    Condition & In plain words\\
    \midrule
    \textcolor{lmpurpled}{\textbf{L1} memory needed} & decisive state hidden when the robot acts\\
    \textcolor{lmpurpled}{\textbf{L2} own evidence} & the change happens in the robot's camera view\\
    \textcolor{lmpurpled}{\textbf{L3} scheduled} & the change starts on a clock, not on agent actions\\
    \textcolor{lmpurpled}{\textbf{L4} visible cause} & a person, robot or machine makes it\\
    \textcolor{lmpurpled}{\textbf{L5} no hints} & no text says whether or when it happens\\
    \textcolor{lmpurpled}{\textbf{L6} no fixed winner} & seeds mix changes, look-alike decoys, none\\
    \textcolor{lmpurpled}{\textbf{L7} controls} & a visible twin keeps the state in view\\
    \textcolor{lmpurpled}{\textbf{L8} solvable} & a privileged script solves every seed\\
    \bottomrule
  \end{tabular*}
  \vspace{-0.6em}
\end{table}

\begin{table*}[t]
  \caption{\textbf{Main results.} Success (\%) of the hidden live-memory families. State changes: event seeds under the
  paused clock (\cP) and the live clock (\cL), and the change \textsf{L}$-$\textsf{P} (\dropc{red}: lower when live).
  None: live success on null seeds, where nobody touches the decisive state. On the realistic tier the first eight
  models run in Claude Code or Codex and the last eight in our own agent with the same tools; on the other two tiers all
  models share one harness. Event seeds per cell: realistic \nrealev, semi-realistic 84, controlled 65 (Qwen3-VL: 62
  paused controlled seeds, three requests exceeded the service's input limit).
  $^\dagger$: open weights. Best live result per column in \textbf{bold}, second \underline{underlined}.}
  \label{tab:main}
  \centering\footnotesize
  \setlength{\tabcolsep}{4.2pt}
  \renewcommand{\arraystretch}{1.0}
  \input{tables/tab_main_body}
  \vspace{-0.4em}
\end{table*}

\subsection{Paused and live episodes}
\label{sec:clock}

\para{World and agent.} An instance is a simulated world with state $s_t$ at world time $t\ge 0$. People, robots and
machines act on it following a script drawn from the seed, and the robot's head camera records frames $o_t$. The agent takes over the robot at time $t_{\text{ho}}$ and works in decisions $i=1,2,\dots$.
Decision $i$ looks at the frames recorded so far, $\mathcal{H}_i=\{o_t: t\le t^o_i\}$, thinks for $\ell_i$ seconds (the
measured time of the model call or agent turn) and returns a command $c_i$: a skill on a numbered mark, \texttt{wait} or
\texttt{done}.

\para{Paused and live clocks.} Paused and live evaluation differ in one thing: how much world time passes while the
model thinks. The command starts at
\begin{equation}
  t^c_i \;=\; t^o_i + \kappa(\ell_i),\qquad \kappa_{\text{paused}}(\ell)=0,\quad \kappa_{\text{live}}(\ell)=\ell .
  \label{eq:clock}
\end{equation}
Under the paused clock the world waits for the model, as in existing benchmarks; under the live clock it keeps running
for as long as the model thinks. To vary latency on its own, the fixed clock $\kappa_L(\ell)=L$ charges every decision
the same $L$ seconds. While the model thinks, the robot holds still, but the other actors keep moving and the camera
keeps recording; these frames reach the model only with its next decision. The command acts on whatever state exists
when it starts, runs for its own duration, and the next decision looks at the world when it ends. The episode ends with
\texttt{done}, an exhausted manipulation budget or a world-time deadline $T$, and succeeds ($R=1$) if the final state
completes the task. The paused and live runs of a seed start from the same state and share the actors' schedule.

\subsection{Decisive variable and live memory}
\label{sec:def}

\para{Decisive variable.} Each instance has a decisive variable $z_t$, a part of the world state, for example which
mark holds Grandpa's mug, whether the tea is already sweetened, or which bin hides the keys. Let $k$ be the decisive
decision (the first manipulation command or \texttt{done}). The episode can succeed only if $c_k$ is right for the value
$z_{t^c_k}$ that holds when the command starts.

\para{Memory.} The instance requires memory if $z$ cannot be read from the current frame when the robot acts: two
reachable states can look the same to the camera and differ in $z$ (identical objects, closed containers, invisible
properties, signals that are over, elapsed time). The agent must infer $z$ from the frames it has recorded.

\para{Live memory.} Another actor (a person, robot or machine) may change $z$ during an interval $[\tau_s,\tau_e]$ after hand-over:
$z_t=z^{-}$ for $t<\tau_s$ and $z_t=z^{+}$ for $t\ge\tau_e$. The instance requires \emph{live memory} if the change can
fall inside the agent's own deliberation, $[\tau_s,\tau_e]\cap[t^o_i,t^c_i)\neq\emptyset$ for some decision $i$. A decision
that observes at $t^o_k<\tau_e$ and commits at $t^c_k\ge\tau_e$ is computed from a record that is true for $z^-$ and
executed in a world with $z^+$. To succeed, the agent must notice that a change is under way, let it finish, and infer
$z^+$ from the frames recorded while it was thinking.

\para{Why a paused clock cannot measure it.} Actors start only while the robot is still (L3 below). Under
$\kappa_{\text{paused}}$ the robot is still only during explicit waits, so a change can begin only when the agent has
chosen to wait, and no change ever begins between an observation and the command computed from it ($t^c_i=t^o_i$).
Paused evaluation tests whether memory is \emph{stored}, and only a running clock tests whether it is kept \emph{true}.

\begin{table*}[t]
  \caption{\textbf{\bench at a glance.} Three tiers that trade realism for control. Families by memory class (U update, W
  write, T time, E execution control). Twins: families with a visible twin. Seeds: per family. Harness: on the realistic tier the model works as a tool-using agent (Claude Code, Codex or ours) that decides when to look, act or wait; on the other tiers it gets one call per step with the frames since its last step. Every instance runs under the paused and the live clock, and visible twins under
  the live clock.}
  \label{tab:tiers}
  \centering\footnotesize
  \setlength{\tabcolsep}{3pt}
  \begin{tabular*}{\textwidth}{@{\extracolsep{\fill}}lllrrrrrrrl rr@{}}
    \toprule
    & & & \multicolumn{5}{c}{Families} & & & & & \\
    \cmidrule(lr){4-8}
    Tier & Simulator & Who changes the state & U & W & T & E & all & Twins & Seeds & Harness & Models & Episodes\\
    \midrule
    Realistic & BEHAVIOR-1K~\cite{li2022behavior1k} & people, robots, machines & 11 & 5 & 0 & 0 & 16 & 9 & 12 & tool-using agent & \nrealmodels & \nrealeps\\
    Semi-realistic & MuJoCo~\cite{todorov2012mujoco} kitchen & a person at the counter & 2 & 3 & 2 & 1 & 8 & 8 & 36 & one call per step & \nsemimodels & \nsemieps\\
    Controlled & ManiSkill~\cite{tao2025maniskill3} & scripted object motions & 3 & 2 & 0 & 0 & 5 & 5 & 40 & one call per step & \nctrlmodels & \nctrleps\\
    \bottomrule
  \end{tabular*}
  \vspace{-0.6em}
\end{table*}

\subsection{Instance conditions}
\label{sec:conditions}

\Cref{tab:conditions} lists eight conditions that make an instance a valid test of live memory. Under L3 the actor
waits until the robot stands still, as a person would before reaching into a shared space. Since the robot is
still while it thinks, live runs place changes inside deliberation. L5 lets the instruction say who shares the workspace
(``your son is setting this table''), identically for all seeds. L6 draws the variant from the seed, so that acting on
the old memory fails event seeds and over-reacting fails decoy and null seeds. We label each family by what the change does to memory: \lmu (\emph{update}: a remembered state is moved,
swapped or hidden), \lmw (\emph{write}: a one-time event or cue must be stored), \lmt (\emph{time}: the answer depends on
elapsed world time), or E (execution control: the clock matters but $z$ stays visible).

\subsection{Evaluation protocol: three explanations of a live drop}
\label{sec:protocol}

\para{Conditions.} Each instance (family and seed) is run in three conditions with the same seed and schedule: hidden
under the paused clock (\cP), hidden under the live clock (\cL), and the visible twin under the live clock (\cV), whose
camera image adds a cue that follows $z$ (a sticker on the target, a lid left ajar, a coloured tag) and changes nothing
else. $S_{\cP}$, $S_{\cL}$ and $S_{\cV}$ denote a model's success rates in the three conditions over a set of
instances, and the live drop is $\Delta=S_{\cP}-S_{\cL}$.

\para{Example.} In \emph{Grandpa's mug} (\cref{fig:teaser}) the robot must bring the mug at mark~1 to Grandpa's tray.
On an event seed the son takes that mug away and puts an identical one in its place, at a time fixed by the seed. In
\cP the world stops while the model thinks, so the swap can never fall between the frame a decision reads and the
moment its command runs. In \cL the swap can happen while the model thinks: its frames arrive only with the next
request, and a ``pick~1'' decided on the earlier frame lifts the look-alike. \cV is \cL with a red sticker on
Grandpa's mug, so the model can find the mug without having followed the swap. \textsf{P}$-$\cL is what the running world
costs, and \textsf{V}$-$\cL is the part of it that comes from having to remember.

A model can lose $\Delta$ for three different reasons, which we define and measure separately.

\para{(i) Latency.} The live clock charges $\ell_i$ to world time. Since $\kappa_{\text{live}}\to\kappa_{\text{paused}}$
as $\ell\to0$, a fast enough agent recovers paused success, and the question is whether the latencies of real models,
seconds, are short enough. We vary latency directly with the fixed clock $\kappa_L$, which charges the same $L$ to every
decision of one model, and compare models whose latencies differ.

\para{(ii) Observation gap.} Agent harnesses observe, think, then act: no frame reaches a decision during
$[t^o_i,t^c_i)$, and the frames of that interval arrive only with the next decision. The decisive decision $k$ is \emph{stale} if its record
predates the end of the change,
\begin{equation}
  \sigma \;=\; \ind\!\left[t^o_k<\tau_e\right],\qquad \lambda \;=\; t^o_k-\tau_e ,
  \label{eq:stale}
\end{equation}
where $\tau_e$ is read from the simulator's privileged log and $\lambda<0$ means the decision was taken before the change
was complete. This gap belongs to the harness rather than to memory, and re-observing the target before acting closes
it. The \emph{commit check} does exactly that at execution time. Let $t^\ell$ be the time of the last frame the agent saw
and $W_m$ a window around the target mark $m$ of command $c_i$. Before $c_i$ runs,
\begin{equation}
  c_i \text{ is cancelled if } \max_{t\in(t^\ell,\,t^c_i]} d\big(W_m(o_t),W_m(o_{t^\ell})\big)>\theta ,
  \label{eq:commit}
\end{equation}
with a neutral message (``not executed: the camera view around mark~$m$ changed since your last observation''), no use of
the budget, and a new decision on the new frames. It needs no model access and no training, and $d$ and $\theta$ are
fixed per tier (\cref{app:harness}). An \emph{offline probe} further asks the same model, without time pressure, a
multiple-choice question about $z_{t^o_k}$ given exactly the images of $\mathcal{H}_k$. A failed episode with a right
probe was lost by acting on the frames, not by misreading them.

\para{(iii) Memory.} The visible twin keeps the change, its timing, the latency and the observation schedule of the
hidden instance and only makes $z$ identifiable at action time. On the families with a twin the live drop therefore
splits into an execution term, which collects latency and the observation gap, and a memory term,
\begin{equation}
  \underbrace{S_{\cP}-S_{\cL}}_{\text{live drop } \Delta} \;=\; \underbrace{(S_{\cP}-S_{\cV})}_{\text{execution}} \;+\;
  \underbrace{(S_{\cV}-S_{\cL})}_{\text{memory } \Delta_{\text{mem}}} .
  \label{eq:decomp}
\end{equation}
With the commit check in place, whatever remains of $\Delta_{\text{mem}}$ is memory that re-observation cannot replace.
For \lmw families the paused clock is not a clean reference (an agent that acts at once never lets world time reach the
cue), so we read $\Delta_{\text{mem}}$ for all classes and $\Delta$ mainly for \lmu. For every difference we compute a
95\% paired bootstrap interval over instances (2{,}000 resamples), reported in \cref{app:results}.

%% file: tables/tab_main_body.tex
  \begin{tabular*}{\textwidth}{@{\extracolsep{\fill}}l rrrr c rrrr c rrrr@{}}
    \toprule
    & \multicolumn{4}{c}{\textbf{Realistic} (BEHAVIOR-1K)} && \multicolumn{4}{c}{\textbf{Semi-realistic} (MuJoCo)} &&
      \multicolumn{4}{c}{\textbf{Controlled} (ManiSkill)}\\
    \cmidrule{2-5}\cmidrule{7-10}\cmidrule{12-15}
    & \multicolumn{3}{c}{state changes} & none && \multicolumn{3}{c}{state changes} & none && \multicolumn{3}{c}{state changes} & none\\
    \cmidrule(lr){2-4}\cmidrule(lr){5-5}\cmidrule(lr){7-9}\cmidrule(lr){10-10}\cmidrule(lr){12-14}\cmidrule(lr){15-15}
    Model & \cP & \cL$\uparrow$ & \textsf{L}$-$\textsf{P} & \cL && \cP & \cL$\uparrow$ & \textsf{L}$-$\textsf{P} & \cL && \cP & \cL$\uparrow$ & \textsf{L}$-$\textsf{P} & \cL\\
    \midrule
    Claude Opus~5.5 & 100.0 & 43.8 & \dropc{$-$56.2} & 92.2 && 97.6 & \underline{98.8} & +1.2 & 100.0 && 89.2 & \textbf{61.5} & \dropc{$-$27.7} & 91.0\\
    GPT-6 Astra & 93.8 & 79.7 & \dropc{$-$14.1} & 98.4 && 100.0 & \textbf{100.0} & 0.0 & 100.0 && 75.4 & \underline{50.8} & \dropc{$-$24.6} & 88.1\\
    Claude Sonnet~5.5 & 98.4 & 34.4 & \dropc{$-$64.1} & 96.9 && 94.0 & 73.8 & \dropc{$-$20.2} & 100.0 && 60.0 & 44.6 & \dropc{$-$15.4} & 77.6\\
    Claude Haiku~5.5 & 95.3 & 21.9 & \dropc{$-$73.4} & 100.0 && 89.3 & 72.6 & \dropc{$-$16.7} & 92.9 && 33.8 & 24.6 & \dropc{$-$9.2} & 77.6\\
    GPT-6 Luna & 95.3 & 7.8 & \dropc{$-$87.5} & 98.4 && 89.3 & 64.3 & \dropc{$-$25.0} & 98.8 && 46.2 & 13.8 & \dropc{$-$32.3} & 67.2\\
    GPT-6.1 Sol & 100.0 & 68.8 & \dropc{$-$31.2} & 98.4 && 100.0 & \underline{98.8} & \dropc{$-$1.2} & 100.0 && 73.8 & 47.7 & \dropc{$-$26.2} & 88.1\\
    GPT-6.1 Terra & 100.0 & 56.2 & \dropc{$-$43.8} & 100.0 && 100.0 & 95.2 & \dropc{$-$4.8} & 100.0 && 61.5 & 49.2 & \dropc{$-$12.3} & 86.6\\
    Claude Fable~5.1 & 93.8 & \textbf{85.9} & \dropc{$-$7.8} & 92.2 && 100.0 & 95.2 & \dropc{$-$4.8} & 100.0 && 73.8 & \underline{50.8} & \dropc{$-$23.1} & 86.6\\
    \midrule
    Gemini~4 Argon & 96.4 & \underline{80.4} & \dropc{$-$16.0} & 93.8 && 100.0 & 88.9 & \dropc{$-$11.1} & 92.4 && 72.3 & 27.7 & \dropc{$-$44.6} & 76.1\\
    Gemini~4 Carbon & 93.8 & 73.4 & \dropc{$-$20.3} & 90.6 && 100.0 & 67.9 & \dropc{$-$32.1} & 90.4 && 98.4 & 6.2 & \dropc{$-$92.3} & 52.2\\
    Grok~4.7 & 96.8 & 34.9 & \dropc{$-$61.9} & 98.4 && 91.7 & 72.6 & \dropc{$-$19.0} & 98.8 && 60.0 & 24.6 & \dropc{$-$35.4} & 49.3\\
    Nova~2 Lite & 88.7 & 14.3 & \dropc{$-$74.4} & 89.1 && 51.2 & 44.0 & \dropc{$-$7.1} & 53.6 && 10.8 & 1.5 & \dropc{$-$9.2} & 25.4\\
    Kimi K2.5$^\dagger$ & 90.5 & 9.5 & \dropc{$-$81.0} & 92.2 && 81.0 & 53.6 & \dropc{$-$27.4} & 89.3 && 23.1 & 9.2 & \dropc{$-$13.8} & 44.8\\
    Qwen3-VL-235B$^\dagger$ & 98.4 & 9.5 & \dropc{$-$88.9} & 98.4 && 64.3 & 26.2 & \dropc{$-$38.1} & 64.3 && 17.7 & 7.7 & \dropc{$-$10.0} & 37.3\\
    Gemma~3~27B$^\dagger$ & 87.3 & 1.6 & \dropc{$-$85.7} & 85.9 && 56.0 & 29.8 & \dropc{$-$26.2} & 54.8 && 21.5 & 7.7 & \dropc{$-$13.8} & 28.4\\
    Ministral~3~14B$^\dagger$ & 76.2 & 12.7 & \dropc{$-$63.5} & 78.1 && 48.8 & 32.1 & \dropc{$-$16.7} & 59.5 && 13.8 & 4.6 & \dropc{$-$9.2} & 19.4\\
    \bottomrule
  \end{tabular*}

%% file: sec/4_benchmark.tex
\section{The \bench Benchmark}
\label{sec:bench}

\bench has 29 task families in three tiers that trade realism for control (\cref{fig:tiers,tab:tiers}). All follow the
protocol of \cref{sec:livemem}, and \cref{app:families} gives each family's decisive variable, event, decoy, null and
visible twin.

\para{Realistic tier.} Sixteen BEHAVIOR-1K scenes in which people, other robots and machines make the change. In the
eleven \lmu scenes a remembered object is swapped with a look-alike (Grandpa's mug, the vegetarian plate under identical
cloches), hidden in a closed container (car keys in lidded bins, a parcel in a locker) or carried off by a machine (a
ball pushed under a sofa by a robot vacuum). In the five \lmw scenes a one-time event decides what remains to be done
(Mom's tea already sweetened, salt in the soup, a ready lamp lit for 6\,s). Decoys reuse the actor and the motion without
changing $z$ (the daughter sweetens the other cup, and the vacuum drives to the ball and back).

\para{Semi-realistic tier.} Eight MuJoCo kitchen families with a person at the counter. \lmu: identical mugs or
covered dishes swapped. \lmw: a hot plate switched on and off before the robot places a plastic bowl (the coil then
looks cold), one of three jars shown once by the cook, and a knife slid under a cover. \lmt: whether the plate was on
for at least 3\,s, and how many sugar cubes dissolved in a tea. An execution control (E) switches the plate on while the robot
decides.

\para{Controlled tier.} Five RoboMME-scene tasks with scripted object motions. \lmu: containers hiding cubes swap three
or four times (SwapStream, InterruptSwap), or identical cubes trade places after a demonstration (RepickStream). \lmw:
targets highlighted once (HighlightLate) or cubes revealed once (RevealLate) after hand-over. Decoys are feints.

\para{Validation.} Before any model runs, every family passes the scripted checks of L3 and L8 (\cref{app:checks}).
A \emph{privileged} policy that reads $z_t$ and waits while an actor is busy solves every realistic seed under both
clocks. A \emph{stale} policy that fixes its plan at the first observation solves every seed paused but fails every
realistic event seed live, and a policy that always assumes a change fails every semi-realistic decoy and null seed.

%% file: sec/5_protocol.tex
\section{Experiments and Findings}
\label{sec:results}

\begin{figure}[t]
  \centering
  \includegraphics[width=\linewidth]{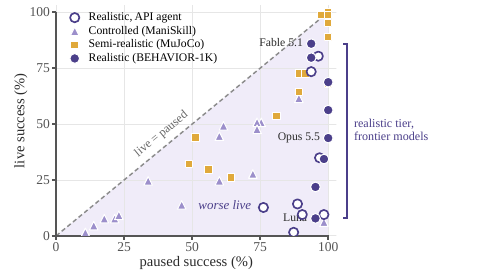}
  \vspace{-1.8em}
  \caption{\textbf{Paused success does not predict live success.} One point per model and tier: success (\%) on the
  event seeds of the hidden families under the paused clock and the live clock. Below the diagonal the model does worse
  live. On the realistic tier, Claude and GPT models with nearly equal paused scores differ widely live. Open
  circles: realistic tier in our own agent.}
  \label{fig:scatter}
  \vspace{-0.8em}
\end{figure}

The three assumptions behind paused evaluation become four questions, each answered by one finding:
\begin{itemize}[leftmargin=2.2em, itemsep=0pt, topsep=2pt, parsep=0pt]
  \item[(Q1)] Does paused success predict live success? (\cref{sec:rq1}, Finding 1)
  \item[(Q2)] Is the live drop a matter of latency, so that a faster agent would avoid it? (\cref{sec:rq2}, Finding 2)
  \item[(Q3)] Is it an observation gap, so that re-observing before acting would avoid it? (\cref{sec:rq3}, Finding 3)
  \item[(Q4)] What remains for memory, and which memory operations break? (\cref{sec:rq4}, Finding 4)
\end{itemize}

\subsection{Setup}
\label{sec:setup}

\para{Models.} We test \nmodels models from \nvendors providers. Eight are Claude and GPT models: Claude Opus~5.5,
Sonnet~5.5, Haiku~5.5 and Fable~5.1, GPT-6 Astra (medium reasoning effort) and Luna (low effort), and GPT-6.1 Sol and
Terra (low effort). On the realistic tier they run inside Claude Code and Codex. The other eight, \extramodels, have no
such agent, so on the realistic tier they run in our own agent, which gives them the same prompt, tools and
observations and charges the world clock only for the time of each model request. On the other two tiers every model
sees the identical prompt and observation (\cref{tab:tiers}). Output limits, effort settings and the models we could
not run are listed in \cref{app:harness}.

\para{Runs and statistics.} Each instance is run once per condition (\cP, \cL, \cV). The commit check is run with Opus
and Astra on the realistic and controlled tiers, and the offline probe for Opus, Sonnet, Haiku, Astra and Luna on all tiers (973
realistic items). Attempts that a service ended for usage limits or rejected as malformed were rerun with the same seed or
excluded, and no number is estimated or imputed. Differences we call reliable have 95\% paired bootstrap intervals over
instances that exclude zero, and all intervals are listed in \cref{app:results}.

%% file: sec/6_results.tex
\begin{figure}[t]
  \centering
  \includegraphics[width=\linewidth]{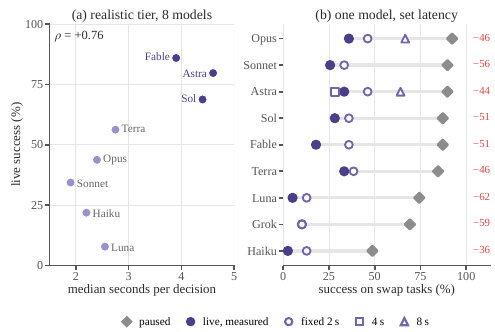}
  \vspace{-1.7em}
  \caption{\textbf{Is it latency?} (a) Realistic tier: median seconds per decision against live success on event seeds,
  one point per model ($\rho$: Spearman). The three slowest models (dark) are the three most robust. (b) Controlled
  swap tasks, event seeds: one model charged a fixed time per decision instead of its measured latency. Red: distance to
  paused success at 2\,s per decision.}
  \label{fig:speed}
  \vspace{-0.8em}
\end{figure}

\begin{figure*}[t]
  \centering
  \includegraphics[width=\textwidth]{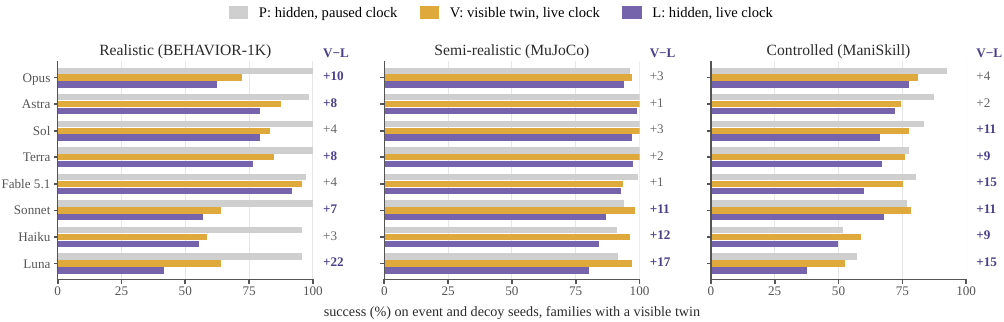}
  \vspace{-1.8em}
  \caption{\textbf{Memory versus execution} (\cref{eq:decomp}). Success (\%) on the families with a visible twin (event
  and decoy seeds) for the hidden family paused (\cP), the visible twin live (\cV) and the hidden family live (\cL).
  \textsf{P}$-$\cV is the cost of acting under the live clock with the change in view (execution), and the number at the right,
  \textsf{V}$-$\cL, is the extra cost of having to remember it (memory, bold if at least 5 points). Having to remember costs
  points on every tier. The same terms with the commit check are in \cref{tab:commit}.}
  \label{fig:decomp}
  \vspace{-0.6em}
\end{figure*}

\subsection{Paused versus live success}
\label{sec:rq1}

\Cref{tab:main,fig:scatter} answer Q1. Of the \npairs model--tier pairs, \nlower lose success under the live clock,
and none gains more than \maxgain points. On the realistic episodes in which the decisive state changes, the Claude and GPT models succeed in \rPevavg\% on average
under the paused clock and in \rLevavg\% under the live clock. Every model succeeds in at least \rPevmin\% paused, and
live the same models succeed in between
\rLevmin\% (\rWorst) and \rLevmax\% (\rBest), so they lose between \rDevmin\ and \rDevmax\ points. The drop
is caused by the changes, not by the clock itself: on null seeds, where nobody touches the decisive state, live success
stays at \rLnullmin--\rLnullmax\% for every model. The ranking also changes. Astra and Fable~5.1 share the lowest paused
score (\rLowPval\%) yet have the two highest live scores, and Opus, perfect when paused, falls to 44\%. The \napi
other models, run in our own agent, show the same pattern: \aPevmin--\aPevmax\% paused against \aLevmin--\aLevmax\% live, and
78--98\% on null seeds. On the
controlled tier all \nctrlmodels models lose \cDmin\ to \cDmax\ points. On the semi-realistic tier, the simplest
scene, the five strongest models lose at most 5 points and the others up to 38.

\begin{finding}{Finding 1 (Q1). Paused success does not predict live success.}
\textbf{No.} When a remembered object changes, the Claude and GPT models fall from \rPevavg\% on average with the world paused to
\rLevavg\% with it running, and a model that never fails when paused succeeds in only 44\%. Every paused score is at
least \rPevmin\%, yet they say nothing about the live ranking. With no change, live success stays at
\rLnullmin--\rLnullmax\%: the loss comes from changes that a paused world never lets happen.
\end{finding}

\begin{figure}[t]
  \centering
  \includegraphics[width=\linewidth]{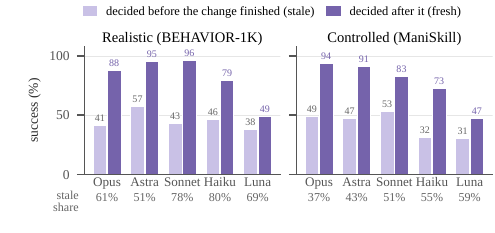}
  \vspace{-1.8em}
  \caption{\textbf{Timing of the decisive decision} (hidden, live, event and decoy seeds). Success (\%) of decisions
  taken on frames from before the other actor's change was complete (stale, $\sigma{=}1$) and after it (fresh). Below
  the axis: share of stale decisions. Stale decisions fail far more often for every model.}
  \label{fig:fresh}
  \vspace{-0.6em}
\end{figure}

\subsection{Latency}
\label{sec:rq2}

\Cref{fig:speed} answers Q2 in two ways. Across models, slower models do not do worse. On the realistic tier the
three slowest models, Astra, Sol and Fable~5.1 (3.9--4.6\,s per decision against 1.9--2.8\,s for the others), are the
three most robust, and median latency and live success on event seeds are positively correlated (Spearman
$\rho=\rhoReal$ over the \nfrontier Claude and GPT models). The two slowest models we test, Gemini~4 Argon and Carbon
(\gArealLat\ and \gCrealLat\,s per decision in our own agent), keep \gArealL\% and \gCrealL\% live. Speed matters only
when the time budget runs out: in the 65\,s controlled episodes Carbon (\gCctrlLat\,s) times out in most live episodes
and reaches \gCctrlL\%. Across the \nctrlmodels models of the other tiers, $\rho=\rhoCtrl$ (controlled) and $\rhoSemi$
(semi-realistic).
Comparisons across models mix speed with capability, so we also set the latency of each model with the fixed clock
$\kappa_L$ on the three swap tasks (\cref{fig:speed}b). Charged 2\,s per decision instead of its measured latency,
eight of nine models gain 5 to 18 points on event seeds (Grok~4.7 none), yet every model stays 36 to 61 points below
its paused success. A slower clock can even help: at 8\,s Opus reaches 67\% and Astra 64\%, against 46\% for both
at 2\,s. The lead time $\lambda$ of \cref{eq:stale} shows why speed alone does not help: a fast decision often falls
inside a change that is still under way. On realistic event seeds Astra's decisive observation comes a median 1.9\,s
\emph{after} the change ends, and those of Opus, Sonnet and Haiku 5.3--8.6\,s before it.

\begin{finding}{Finding 2 (Q2). A faster agent would not avoid the drop.}
\textbf{No.} Slower models do no worse, and even at 2\,s per decision every model tested stays at least 36 points below its paused
success. Speed helps only if the decision waits for the change to finish.
\end{finding}

\subsection{Observation gap}
\label{sec:rq3}

\para{Acting on old frames.} \Cref{fig:fresh} splits the live episodes by $\sigma$, \ie by whether the decision
that settles the task was made on frames from before the change ended (stale) or after it (fresh). On the realistic
tier 51--80\% of these decisions are stale, and they succeed in 38--57\% of episodes, against 79--96\% for fresh
decisions (49\% for Luna). On event seeds, Opus succeeds in 1 of 34 stale and 21 of 23 fresh episodes. The controlled tier shows the same split
(31--53\% against 47--94\%).

\para{The frames are read correctly.} The offline probe (\cref{tab:probe}) asks each model about the decisive state
using exactly the frames it had. Opus, Sonnet and Haiku answer correctly in 76--89\% of their failed realistic
episodes, and almost all of those are stale decisions. Opus and Astra show the same profile on the controlled tier (29
of 30 and 30 of 37). Rarer failure modes of the weaker models: \cref{fig:fail}.

\begin{table}[t]
  \caption{\textbf{Commit check} (\cref{eq:commit}). Success (\%) under the live clock on event and decoy seeds, without
  (min.) and with the check, and the memory term $\Delta_{\text{mem}}=S_{\cV}-S_{\cL}$ with the check. Realistic: the 13
  scenes where the check applies (hidden $n{=}104$, twins $n{=}56$). Controlled: the three swap tasks ($n{=}80$).
  Re-observing closes the realistic gap but not the controlled one.}
  \label{tab:commit}
  \centering\footnotesize
  \begin{tabular*}{\columnwidth}{@{\extracolsep{\fill}}ll rr rr r@{}}
    \toprule
    & & \multicolumn{2}{c}{Hidden (\cL)} & \multicolumn{2}{c}{Visible twin (\cV)} & \\
    \cmidrule(lr){3-4}\cmidrule(lr){5-6}
    Tier & Model & min. & check & min. & check & $\Delta_{\text{mem}}$\\
    \midrule
    \input{tables/tab_commit_body}
  \end{tabular*}
  \vspace{-0.8em}
\end{table}

\para{Re-observing before acting.} If stale commitment is an observation gap, closing the gap should help. The
commit check does this simply: just before a command runs, the robot looks at the target once more, and if it has
changed, the agent decides again on the new frames.
\Cref{tab:commit} confirms it. On the 13 realistic scenes where the commit check applies, hidden live success rises to
88--97\% for all five models tested (Opus from 62\% to 97\%, Astra from 81\% to 95\%), the memory term stays within
4 points of zero, and for Opus and Astra nearly every executed decisive command becomes fresh. On the controlled swap tasks the check
raises hidden live success by 9 to 21 points for the five strongest models, but not for Haiku, Fable~5.1 and Grok~4.7
(3 to 13 points lower).

\begin{finding}{Finding 3 (Q3). Agents see the change but act on old frames.}
\textbf{Yes.} Most failures are decisions taken on frames recorded before the other actor finished, by models that read those frames
correctly. Looking once more before acting recovers most of the realistic gap, with no training.
\end{finding}

\subsection{Memory}
\label{sec:rq4}

\para{Memory versus execution.} \Cref{fig:decomp} applies \cref{eq:decomp} to the families with a visible twin. With
the decisive state kept in view, the same change at the same time costs less on every tier. $\Delta_{\text{mem}}$ is
positive for all eight models on all three tiers, from +1 to +22 points. On the controlled swap tasks the commit check (\cref{tab:commit}) lifts the twins of the five strongest models to
95--100\% but their hidden tasks only to 68--81\%: when identical objects are swapped several times, cancelling a
command forces a new decision but cannot tell the agent which object is which. On the realistic tier the remembered
object changes once, so one more look settles it; on the controlled swaps it changes several times, so the agent must
have followed every swap. The memory term therefore
\emph{grows} with the check, to 15--31 points against 4--18 without it, and with the check it is positive for all
nine models tested (12--58 points, every 95\% interval above zero; \cref{tab:app_commitci}).

\para{Updates break first.} By memory class (\cref{fig:blocks}), updates lose the most. On the realistic \lmu
scenes, where a remembered object is swapped, hidden or carried away, live success is 75\% for Astra, 25\% for Opus and
2--18\% for Sonnet, Haiku and Luna, against 91--100\% paused; on the controlled swaps Opus and Astra fall from about
90\% to about 35\%. Writes are solved live by the stronger models on every tier.

\begin{finding}{Finding 4 (Q4). Memory still fails on repeated swaps.}
\textbf{Updates.} Having to remember a change rather than see it costs every model points on every tier (up to 22). When identical
objects are swapped several times, looking again does not help: all nine models tested still lose at least 12 points.
Keeping track of which object is which while the world moves is the core of live memory, and \bench isolates it.
\end{finding}

%% file: tables/tab_commit_body.tex
    Realistic & Opus & 61.5 & 97.1 & 64.3 & 100.0 & 0.0\\
     & Astra & 80.8 & 95.2 & 83.9 & 100.0 & 1.8\\
     & Sol & 77.9 & 92.3 & 78.6 & 100.0 & 1.8\\
     & Terra & 68.3 & 87.5 & 80.4 & 92.9 & $-$1.8\\
     & Fable~5.1 & 86.5 & 96.2 & 94.6 & 100.0 & 3.6\\
    \midrule
    Controlled & Opus & 62.5 & 80.0 & 68.8 & 95.0 & 15.0\\
    swaps & Astra & 61.2 & 81.2 & 65.0 & 100.0 & 18.8\\
     & Sonnet & 58.8 & 67.5 & 73.8 & 98.8 & 31.2\\
     & Haiku & 40.0 & 37.5 & 58.8 & 72.5 & 35.0\\
     & Luna & 33.8 & 41.2 & 52.5 & 81.2 & 40.0\\
     & Sol & 51.2 & 72.5 & 68.8 & 98.8 & 26.3\\
     & Terra & 53.8 & 73.8 & 68.8 & 100.0 & 26.2\\
     & Fable~5.1 & 38.8 & 26.2 & 68.8 & 83.8 & 57.5\\
     & Grok~4.7 & 23.8 & 20.0 & 53.8 & 32.5 & 12.5\\
    \bottomrule

%% file: sec/7_conclusion.tex
\section{Conclusion}
\label{sec:conclusion}

We built \bench to ask whether robot memory is solved when the world does not wait. Frontier models that succeed
in \rPevavg\% of our changing episodes on average when the world is paused succeed in only \rLevavg\% once it keeps
moving, mostly because they decide before a change they are watching is over. Looking again before acting recovers most
of this gap; what remains, tracking identical objects through repeated swaps, is the memory problem that robots in
shared spaces still have to solve.

\para{Limitations.} All scenes are simulated, other actors follow scripts, the people in the realistic tier have no
articulated arms, and latency depends on the load of public model services. Realistic-tier results should be
compared within models that run in the same kind of agent.

%% file: sec/X_suppl.tex
\clearpage
\setcounter{page}{1}
\maketitlesupplementary
\appendix

\section{Task Families}
\label{app:families}

\Cref{tab:app_real,tab:app_semi} describe every family. All variants of a family start from the same scene; the
variant is drawn from the seed (event, decoy, null in turn). Instructions may say who shares the workspace (``Your son is
setting this table this morning'') identically for all seeds, and never whether or when anything happens.

\begin{table*}[t]
  \caption{\textbf{Realistic tier (BEHAVIOR-1K).} What changes in each scene. U/W: \lmu/\lmw. On null seeds the actor
  comes and goes, or the machine idles, without changing anything. Twin: the cue that keeps the state in view (--: none).}
  \label{tab:app_real}
  \centering\scriptsize
  \setlength{\tabcolsep}{3pt}
  \renewcommand{\arraystretch}{1.08}
  \begin{tabularx}{\textwidth}{@{}l c l X l l@{}}
    \toprule
    {\tableheader Scene} & & {\tableheader Actor} & {\tableheader Event (the state to remember)} & {\tableheader Decoy} &
      {\tableheader Twin}\\
    \midrule
    Grandpa's mug & U & son & the medicine mug (1 of 3 identical) moved away, a look-alike put in its place & only a look-alike moved & red sticker\\
    Express tote & U & picker & the express tote (1 of 3 identical) moved away, a look-alike put in its place & only a look-alike moved & red sticker\\
    Car keys & U & teenager & keys put into 1 of 3 lidded baskets & put in one, later moved & last lid ajar\\
    Gauge in bins & U & technician & gauge put into 1 of 3 lidded bins & put in one, later moved & last lid ajar\\
    Mom's tea & W & daughter & Mom's tea gets its spoon of sugar & the other cup sweetened & token at cup\\
    Catalyst dose & W & technician & flask A gets its dose & flask B dosed & token at flask\\
    Put away & U & helper robot & apple put into closed bin $k$ & bin $k$ opened, apple returned & lid left off\\
    Anna's mug & U & helper robot & Anna's mug (1 of 2 identical) moved, a board put in its place & the other mug moved & --\\
    Vacuum and ball & U & robot vacuum & ball pushed under sofa $k$ & drives to the ball and back & --\\
    Salted soup & W & helper robot & soup salted & same motion, pepper mill & --\\
    Rotary reject & U & rotary table & table indexes 1--5 steps with the reject in a nest & full turn & --\\
    Conveyor lanes & U & conveyor & order tote routed to 1 of 3 lanes & tote re-scanned, out and back & --\\
    Press lamp & W & press & cycle ends: lamp lit for 6\,s & vents twice, no lamp & lamp stays lit\\
    Parcel locker & U & courier robot & parcel stored in identical locker $k$, door closed & locker $k$ opened and closed empty & door left open\\
    Kitting partner & W & partner robot & 1--3 covered totes kitted & same lids opened and closed & --\\
    Cloche pass & U & expo robot & two cloches moved, one over the vegetarian plate & the other two moved & --\\
    \bottomrule
  \end{tabularx}
\end{table*}

\begin{table*}[t]
  \caption{\textbf{Semi-realistic (MuJoCo kitchen) and controlled (ManiSkill) families.} On null seeds the hand passes
  far (semi-realistic) or nothing moves (controlled); in Shown jar, HighlightLate and RevealLate the cue is then given only
  at hand-over. T: \lmt; E: execution control.}
  \label{tab:app_semi}
  \centering\scriptsize
  \setlength{\tabcolsep}{3pt}
  \renewcommand{\arraystretch}{1.08}
  \begin{tabularx}{\textwidth}{@{}l c X l l@{}}
    \toprule
    {\tableheader Family} & & {\tableheader Event} & {\tableheader Decoy} & {\tableheader Twin}\\
    \midrule
    \multicolumn{5}{@{}l}{\emph{Semi-realistic: a person at the counter; UR5e with pick, place and press on marked places}}\\
    Mugs & U & two identical mugs swapped & both nudged, put back & target mug red\\
    Covered dishes & U & two covered dishes swapped while the robot holds the salt & other cover lifted, put back & apple's cover blue\\
    Residual heat & W & plate switched on, then off, before the robot places a plastic bowl & hand reaches, never presses & plate stays on\\
    Shown jar & W & cook's hand over 1 of 3 jars once, for 1\,s, after hand-over & shown at hand-over, later a sweep & hand stays\\
    Knife & W & knife slid under 1 of 2 covers & knife lifted, put back & knife moved in view\\
    Heat duration & T & plate on for 4\,s (hot after 3\,s) & on for 1\,s & plate stays on\\
    Sugar count & T & 2 cubes dropped into the tea (enough) & 1 cube (not enough) & cubes on the saucer\\
    Heat (E) & E & plate switched on while the robot decides where to place the bowl & reaches twice, never presses & plate stays on\\
    \midrule
    \multicolumn{5}{@{}l}{\emph{Controlled: scripted motions on RoboMME scenes; Franka Panda with RoboMME's skills; 20\,Hz control}}\\
    SwapStream & U & containers hiding cubes swap 3 or 4 times & as many feints (move 35\%, return) & tag on each cube's container\\
    InterruptSwap & U & as SwapStream, on the InterruptSwap scenes & feints & tags\\
    RepickStream & U & identical cubes swap after the video that showed which to re-pick & feints & tag on the picked cube\\
    HighlightLate & W & targets highlighted once after hand-over & at hand-over, repeated later & tag from the highlight\\
    RevealLate & W & containers lift once after hand-over, showing the cubes & at hand-over, repeated later & tags from the reveal\\
    \bottomrule
  \end{tabularx}
\end{table*}

\section{Clock-Sensitivity Checks}
\label{app:checks}

\begin{table}[t]
  \caption{\textbf{Formal statement of the instance conditions} of \cref{tab:conditions}. The controlled tier relaxes L4
  by design; the checks are described below.}
  \label{tab:app_conditions}
  \centering\footnotesize
  \setlength{\tabcolsep}{3pt}
  \renewcommand{\arraystretch}{1.1}
  \begin{tabularx}{\linewidth}{@{}l X@{}}
    \toprule
    {\tableheader Condition} & {\tableheader Formal statement}\\
    \midrule
    L1 memory needed & $z_{t^c}$ cannot be read from $o_{t^c}$\\
    L2 own evidence & the change is a continuous motion in $\{o_t\}_{t\in[\tau_s,\tau_e]}$\\
    L3 scheduled & $\tau_s$: first idle time after $t_{\text{ho}}+u$, $u\sim U(0.3,2.5)$\,s, independent of the agent\\
    L4 visible cause & a person, robot or machine with a reason to act\\
    L5 no hints & no text says whether or when anything happens\\
    L6 no fixed winner & event $z^+\!\neq\!z^-$; decoy: motion with $z^+\!=\!z^-$; null: no motion\\
    L7 controls & a visible twin keeps $z_t$ identifiable\\
    L8 solvable & a privileged policy solves every seed under both clocks\\
    \bottomrule
  \end{tabularx}
  \vspace{-0.6em}
\end{table}

\Cref{tab:app_conditions} states the conditions every instance must meet. Before any model is run, every family
passes a scripted check with three policies: a \emph{privileged} policy that
reads the true state and waits while another actor is busy; a \emph{stale} policy that fixes its plan from the first
observation, charges 3\,s per decision and never re-checks; and, on the semi-realistic tier, an \emph{always-change}
policy that assumes the change happened. A family passes if the privileged policy succeeds under both clocks, the stale
policy succeeds paused and fails live on event seeds while passing decoy and null seeds, and fixed policies cannot win.

\para{Realistic.} 528 scripted episodes, 16 hidden scenes and 9 visible twins: the privileged policy solves every
seed under both clocks; the stale policy solves every seed paused and, live, fails every event seed and passes every
decoy and null seed, except the two-step decoys of \emph{Car keys} and \emph{Gauge in bins}, which hide the item twice and
are real changes for a stale policy.

\para{Semi-realistic.} Stale policy at 3\,s per decision, event seeds: mugs 0/2, covered dishes 0/2, residual heat 0/2,
knife 1/2 and shown jar 1/2 (a random guess among two covers or three jars), heat duration 0/4, sugar count 0/4; all
paused, decoy and null seeds succeed. The latency-aware privileged policy succeeds on every seed under both clocks. The
always-change policy (run on mugs, covered dishes and heat) fails every decoy and null seed.

\para{Controlled.} Episodes 0--39 per task. Stale policy at 3\,s, event seeds: SwapStream 6/13, RepickStream 3/13,
InterruptSwap 9/13 (a swap may spare the target); a frame-blind policy on HighlightLate and RevealLate 0/13; decoy and
null seeds pass. The latency-aware policy solves every event seed of SwapStream and InterruptSwap at 3\,s.

\begin{figure*}[t]
  \centering
  \includegraphics[width=\textwidth]{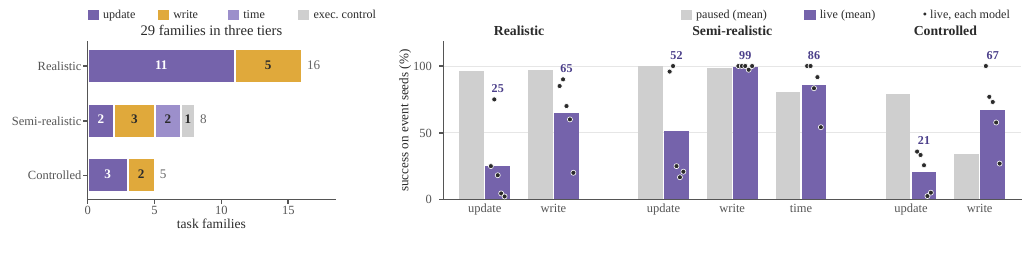}
  \vspace{-1.6em}
  \caption{\textbf{Which memory operations break.} Left: the 29 families by tier and memory class. Right: success (\%)
  on the event seeds of each tier and class, paused and live, for the five frontier models (bars: mean, dots: models).
  Updates (U) lose the most on the realistic and controlled tiers.}
  \label{fig:blocks}
  \vspace{-0.6em}
\end{figure*}

\section{Harnesses and Settings}
\label{app:harness}

\begin{table*}[t]
  \caption{\textbf{Harness settings.} Top: agent and budget. Bottom: output-token limit per model (default: the agent's
  own). Gemini runs through the Gemini API and every other API model through Amazon Bedrock.}
  \label{tab:app_harness}
  \centering\scriptsize
  \setlength{\tabcolsep}{4pt}
  \renewcommand{\arraystretch}{1.08}
  \begin{tabularx}{\textwidth}{@{}l X X@{}}
    \toprule
    & {\tableheader Realistic} & {\tableheader Semi-realistic / controlled}\\
    \midrule
    Agent & Claude Code 2.1.314 or Codex with MCP robot tools; other models: our API agent, same tools as JSON calls & one API call per decision\\
    Frames & current view with marks, contact sheet since the last request, any window on request & $\le$30 frames since the last decision (2\,Hz), newest with marks (controlled: also unmarked; demo frames on turn 1)\\
    History & the agent's own context (API agent: last 12 images) & semi: text log of commands and notes; controlled: conversation, older images replaced by a placeholder\\
    Budget & 2--3 manipulation actions, 300\,s & 240\,s, 12 decisions / 65\,s, 25 decisions\\
    \midrule
    Opus, Fable~5.1 & default & Opus 4{,}096; Fable~5.1 16k (semi), 4{,}096 (controlled)\\
    Sonnet, Haiku & 32k & 4{,}096\\
    GPT (Codex) & default; Astra medium effort, Luna, Sol, Terra low & Astra 16k; Luna 16--20k; Sol, Terra 16k; same effort\\
    Gemini~4 & 16k & 16k\\
    Grok~4.7 & 8k & 16k (semi), 8k (controlled)\\
    Kimi K2.5 & 4{,}096 & 4{,}096 / 8k\\
    Qwen3-VL, Gemma~3, Nova~2 Lite, Ministral~3 & 4{,}096 & 4{,}096\\
    \bottomrule
  \end{tabularx}
\end{table*}

\Cref{tab:app_harness} lists the agent, the budget and the output-token limit of every model.

\para{Models not evaluated.} Kimi K3 was not evaluated because its service rejected nearly every request for
quota, Llama~4 Maverick and Mistral Large~3 accept at most three images per request, and GLM~5.3 accepts none.

\para{Prompt.} Each realistic prompt has the task, the clock rules and the tool list. For \emph{Grandpa's mug},
live clock:
\begin{quote}\scriptsize\ttfamily
TASK (Grandpa's mug): Your robot has just spent the last 10 seconds sorting cutlery on the side cart; its head camera
kept recording the whole time, and your first observation includes that recording. When your robot started wiping,
Grandpa's mug stood at mark 1. The three mugs look identical, but Grandpa's has his heart medicine dissolved in it. Bring
Grandpa's mug to his tray (mark 6). Your son is setting this table this morning. You have 2 manipulation actions (pick,
place).\\[2pt]
CLOCK: LIVE clock: the world keeps running while you think. Other robots, machines and people keep moving. Every response
includes the robot camera frames recorded since your previous request, so you can see what happened while you were
thinking. Your robot never reaches into a place another robot, machine or person is handling right now: such a command
first waits until they are done (world time runs).\\[2pt]
Tools: robot\_spec, robot\_observe, robot\_frames(t\_from, t\_to), robot\_act(skill, mark), robot\_wait(seconds),
robot\_done. [\dots] A skill acts on whatever is at the numbered place when it runs, not on what was there when you last
looked.
\end{quote}
The paused prompt replaces the clock paragraph with its paused counterpart. The agent runs in an empty working directory
with a random name, so neither the scene nor the seed can be read from the environment.

\para{Commit check.} Realistic tier: a $128{\times}128$ window around the target mark; after a $5{\times}5$ blur, a
change is more than 2\% of pixels whose grey level differs by more than 24. Controlled tier: at least 20 pixels in the
target region whose largest channel difference exceeds 40. A cancelled command uses no manipulation budget.

\para{Probe.} Each item gives the model the images it received before its decisive decision, in order, with the tool
text, and a multiple-choice question about the decisive state at that time; the answer comes from the simulator's
privileged log. The realistic tier has 973 items, all answered.

\section{Additional Results}
\label{app:results}

\begin{table}[t]
  \caption{\textbf{Live drop and its two terms} for the eight Claude and GPT models, in points: $\Delta=S_{\cP}-S_{\cL}$
  (\dropc{red}) over all seeds of the hidden live-memory families; on the families with a visible twin (event and decoy
  seeds) the execution term $S_{\cP}-S_{\cV}$ and the memory term $S_{\cV}-S_{\cL}$. Bold: the 95\% paired bootstrap
  interval over instances (2{,}000 resamples) excludes zero; the intervals are in \cref{fig:ci}. Last column: median
  seconds per decision (hidden, live). With the commit check: \cref{tab:app_commitci}.}
  \label{tab:app_decomp}
  \centering\footnotesize
  \setlength{\tabcolsep}{4.5pt}
  \begin{tabular}{@{}c@{\hspace{4pt}}l rrrr@{}}
    \toprule
    & Model & $\Delta$ & \textsf{P}$-$\cV & \textsf{V}$-$\cL & s/dec.\\
    \midrule
\input{tables/tab_app_ci_frontier_body}
  \end{tabular}
  \vspace{-0.6em}
\end{table}

\begin{table}[t]
  \caption{\textbf{Memory term with the commit check}, $S_{\cV}-S_{\cL}$ in points on event and decoy seeds (realistic: the
  13 scenes where the check applies; controlled: the three swap tasks), with the bounds of its 95\% paired bootstrap
  interval over instances.}
  \label{tab:app_commitci}
  \centering\footnotesize
  \setlength{\tabcolsep}{4pt}
  \begin{tabular}{@{}ll rrr@{}}
    \toprule
    Tier & Model & $\Delta_{\text{mem}}$ & lower & upper\\
    \midrule
\input{tables/tab_app_commitci_body}
  \end{tabular}
\end{table}

\begin{table*}[t]
  \caption{\textbf{Update and write families per tier} (hidden, event seeds). Success (\%) under the paused clock
  (\cP) and the live clock (\cL). --: not run.}
  \label{tab:app_class}
  \centering\footnotesize
  \setlength{\tabcolsep}{3.6pt}
  \begin{tabular}{@{}l rrrr c rrrr c rrrr@{}}
    \toprule
    & \multicolumn{4}{c}{Realistic} && \multicolumn{4}{c}{Semi-realistic} && \multicolumn{4}{c}{Controlled}\\
    \cmidrule{2-5}\cmidrule{7-10}\cmidrule{12-15}
    & \multicolumn{2}{c}{Update (\lmu)} & \multicolumn{2}{c}{Write (\lmw)} && \multicolumn{2}{c}{Update (\lmu)} &
      \multicolumn{2}{c}{Write (\lmw)} && \multicolumn{2}{c}{Update (\lmu)} & \multicolumn{2}{c}{Write (\lmw)}\\
    \cmidrule(lr){2-3}\cmidrule(lr){4-5}\cmidrule(lr){7-8}\cmidrule(lr){9-10}\cmidrule(lr){12-13}\cmidrule(lr){14-15}
    Model & \cP & \cL & \cP & \cL && \cP & \cL & \cP & \cL && \cP & \cL & \cP & \cL\\
    \midrule
\input{tables/tab_app_class_body}
  \end{tabular}
\end{table*}

\begin{figure}[t]
  \centering
  \includegraphics[width=\linewidth]{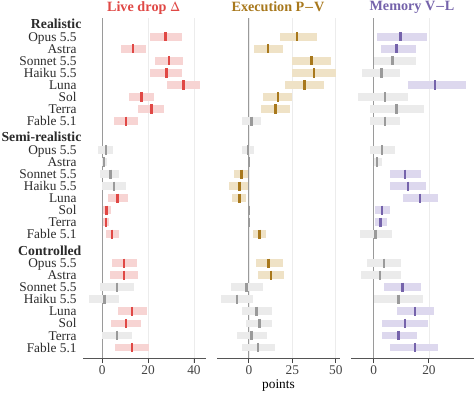}
  \vspace{-1.6em}
  \caption{\textbf{95\% paired bootstrap intervals of \cref{tab:app_decomp}} over instances (2{,}000 resamples), in
  points. Bar: the interval; tick: the estimate. Gray: the interval includes zero.}
  \label{fig:ci}
\end{figure}

\para{Live drop and its two terms.} \Cref{tab:app_decomp} gives the live drop of the eight Claude and GPT models with
its execution and memory terms, \cref{fig:ci} their 95\% intervals, and \cref{tab:app_commitci} the memory term with
the commit check.

\para{Memory classes.} \Cref{fig:blocks} groups the families by memory class: updates (U) lose the most on the realistic and
controlled tiers.
\Cref{tab:app_class} shows the update drop on every tier, including Opus and Astra on the controlled swaps (92\% and
90\% paused, 36\% and 33\% live). The controlled write tasks illustrate why we read \lmw through the twin: paused, an agent
that acts at once never lets the highlight or reveal appear (Sonnet 15\%), while live it often does (73\%).

\begin{table}[t]
  \caption{\textbf{Offline probe on failed live episodes} (hidden, event and decoy seeds). Right: given the same frames
  without time pressure, the model answers the decisive question correctly. Stale: of those, episodes decided before
  the change was complete.}
  \label{tab:probe}
  \centering\footnotesize
  \setlength{\tabcolsep}{3.2pt}
  \begin{tabular}{@{}l rrrr c rrrr@{}}
    \toprule
    & \multicolumn{4}{c}{Realistic} && \multicolumn{4}{c}{Controlled}\\
    \cmidrule{2-5}\cmidrule{7-10}
    Model & fails & right & \% & stale && fails & right & \% & stale\\
    \midrule
    Opus~5.5   & 50 & 38 & 76 & 37 && 30 & 29 & 97 & 24\\
    Astra      & 29 & 16 & 55 & 13 && 37 & 30 & 81 & 24\\
    Sonnet~5.5 & 56 & 50 & 89 & 50 && 43 & 26 & 60 & 21\\
    Haiku~5.5  & 57 & 46 & 81 & 46 && 66 & 28 & 42 & 22\\
    Luna       & 69 & 30 & 43 & 23 && 80 & 32 & 40 & 21\\
    \bottomrule
  \end{tabular}
  \vspace{-0.8em}
\end{table}

\begin{table}[t]
  \caption{\textbf{Decision timing, all tiers} (hidden live, event and decoy seeds): share (\%) of stale decisions and success
  (\%) of stale (s) and fresh (f) decisions.}
  \label{tab:app_fresh}
  \centering\scriptsize
  \setlength{\tabcolsep}{3pt}
  \begin{tabular}{@{}l rrr rrr rrr@{}}
    \toprule
    & \multicolumn{3}{c}{Realistic} & \multicolumn{3}{c}{Semi} & \multicolumn{3}{c}{Controlled}\\
    \cmidrule(lr){2-4}\cmidrule(lr){5-7}\cmidrule(lr){8-10}
    Model & stale & succ.\,s & succ.\,f & stale & succ.\,s & succ.\,f & stale & succ.\,s & succ.\,f\\
    \midrule
    Opus   & 61 & 41 & 88 & 12 & 100 & 93 & 37 & 49 & 94\\
    Astra  & 51 & 57 & 95 & 13 & 100 & 98 & 43 & 47 & 91\\
    Sonnet & 78 & 43 & 96 & 13 & 100 & 80 & 51 & 53 & 83\\
    Haiku  & 80 & 46 & 79 & 3  & 100 & 81 & 55 & 32 & 73\\
    Luna   & 69 & 38 & 49 & 12 & 86  & 69 & 59 & 31 & 47\\
    \bottomrule
  \end{tabular}
\end{table}

On the semi-realistic tier few decisions are stale (\cref{tab:app_fresh}), and the smaller models' failures are fresh decisions: they had the
finished change in their frames and did not use it, which matches the probe (\cref{tab:probe}) and the memory term of
\cref{fig:decomp}.

\begin{figure}[t]
  \centering
  \includegraphics[width=\linewidth]{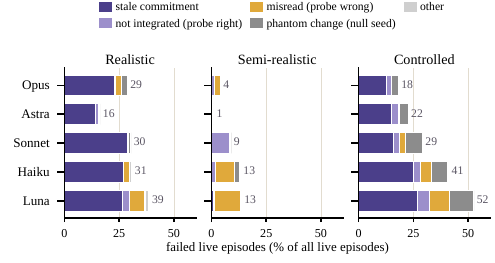}
  \vspace{-1.8em}
  \caption{\textbf{Failure taxonomy} of live hidden episodes (all seeds), five frontier models. Stale: decided before the
  change was complete. Not integrated and misread: fresh decision with the probe right and wrong. Phantom: failure on a
  null seed. Stale commitment dominates the realistic and controlled tiers.}
  \label{fig:fail}
\end{figure}

\para{Failure modes.} \Cref{fig:fail} combines timing and probe into a taxonomy. Stale commitment dominates the
realistic and controlled tiers for every model. On the kitchen tier failures are fresh. Sonnet saw the finished change
(probe right in 21 of 22 failures, none stale) and still acted on the old memory, a failure to \emph{integrate} the
update, while Haiku and Luna \emph{misread} it (probe wrong). Phantom changes, failures on no-change seeds, are rare on
the realistic tier (at most 5 of 64 null seeds) and more frequent on the controlled tier.

\begin{figure}[t]
  \centering
  \includegraphics[width=\linewidth]{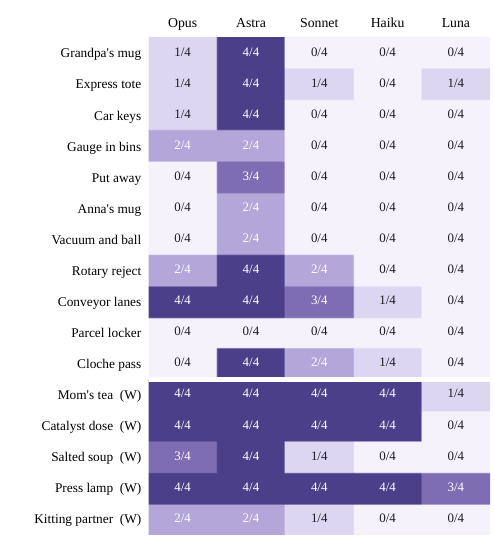}
  \vspace{-1.6em}
  \caption{\textbf{Per-scene live success on the realistic tier} (hidden, event seeds; successes of 4). Paused, every
  cell is 3/4 or 4/4 except Astra on Rotary reject (1/4). W: write scenes; the others are updates.}
  \label{fig:scenes}
\end{figure}

\para{Per scene.} \Cref{fig:scenes} shows that the update scenes separate the models most: Astra solves 10 of the 11
update scenes at least half of the time live, Opus and Sonnet 3, Haiku and Luna none. Parcel locker, where a courier
robot puts the parcel into one of six identical lockers and closes the door, defeats every model live.

\para{Data integrity.} Every number comes from a completed episode; nothing is estimated or imputed. Attempts that
ended because a model service refused requests for usage limits (25 on the realistic tier) were rerun with the same
seed and excluded. Before the main run we fixed a renderer lag that could hide the last discrete change from the head
camera and moved two marks that occluded the decisive object; all affected episodes were discarded and rerun. The
visible twins of the two controlled cue tasks first carried a sentence explaining their tags; their minimal-harness runs
were repeated without it and only those are reported. Their commit-harness runs keep the sentence, which is why
\cref{tab:commit} reports the controlled commit check on the three swap tasks only.

%% file: tables/tab_app_ci_frontier_body.tex
    \multirow{8}{*}{\rotatebox{90}{\scriptsize Realistic}} & Opus~5.5 & \dropc{\textbf{+27.6}} & \textbf{+27.8} & \textbf{+9.7} & 2.4\\
     & Astra & \dropc{\textbf{+13.5}} & \textbf{+11.1} & \textbf{+8.3} & 4.6\\
     & Sonnet~5.5 & \dropc{\textbf{+29.2}} & \textbf{+36.1} & +6.9 & 1.9\\
     & Haiku~5.5 & \dropc{\textbf{+28.1}} & \textbf{+37.5} & +2.8 & 2.2\\
     & Luna & \dropc{\textbf{+35.4}} & \textbf{+31.9} & \textbf{+22.2} & 2.6\\
     & Sol & \dropc{\textbf{+17.2}} & \textbf{+16.7} & +4.2 & 4.4\\
     & Terra & \dropc{\textbf{+21.4}} & \textbf{+15.3} & +8.3 & 2.8\\
     & Fable~5.1 & \dropc{\textbf{+10.4}} & +1.4 & +4.2 & 3.9\\
    \midrule
    \multirow{8}{*}{\rotatebox{90}{\scriptsize Semi-real.}} & Opus~5.5 & \dropc{+1.6} & $-$0.6 & +3.0 & 4.2\\
     & Astra & \dropc{+0.8} & +0.0 & +1.2 & 2.6\\
     & Sonnet~5.5 & \dropc{+3.6} & \textbf{$-$4.2} & \textbf{+11.3} & 2.6\\
     & Haiku~5.5 & \dropc{+5.2} & \textbf{$-$5.4} & \textbf{+12.5} & 3.1\\
     & Luna & \dropc{\textbf{+6.7}} & \textbf{$-$5.4} & \textbf{+16.7} & 2.1\\
     & Sol & \dropc{\textbf{+2.0}} & +0.0 & \textbf{+3.0} & 2.7\\
     & Terra & \dropc{\textbf{+1.6}} & +0.0 & \textbf{+2.4} & 2.1\\
     & Fable~5.1 & \dropc{\textbf{+4.4}} & \textbf{+6.0} & +0.6 & 6.3\\
    \midrule
    \multirow{8}{*}{\rotatebox{90}{\scriptsize Controlled}} & Opus~5.5 & \dropc{\textbf{+9.5}} & \textbf{+11.3} & +3.8 & 4.2\\
     & Astra & \dropc{\textbf{+9.5}} & \textbf{+12.8} & +2.3 & 2.6\\
     & Sonnet~5.5 & \dropc{+6.5} & $-$1.5 & \textbf{+10.5} & 3.1\\
     & Haiku~5.5 & \dropc{+1.0} & $-$6.8 & +9.0 & 3.6\\
     & Luna & \dropc{\textbf{+13.0}} & +4.5 & \textbf{+15.0} & 2.4\\
     & Sol & \dropc{\textbf{+10.5}} & +6.0 & \textbf{+11.3} & 2.7\\
     & Terra & \dropc{+6.5} & +1.5 & \textbf{+9.0} & 2.2\\
     & Fable~5.1 & \dropc{\textbf{+13.0}} & +5.3 & \textbf{+15.0} & 5.9\\
    \bottomrule

%% file: tables/tab_app_commitci_body.tex
    Realistic & Opus & 0.0 & 0.0 & 0.0\\
     & Astra & 1.8 & 0.0 & 5.4\\
     & Sol & 1.8 & 0.0 & 5.4\\
     & Terra & $-$1.8 & $-$7.1 & 3.6\\
     & Fable~5.1 & 3.6 & 0.0 & 8.9\\
    \midrule
    Controlled swaps & Opus & 15.0 & 6.2 & 25.0\\
     & Astra & 18.8 & 11.3 & 27.5\\
     & Sonnet & 31.2 & 21.3 & 41.2\\
     & Haiku & 35.0 & 22.5 & 47.5\\
     & Luna & 40.0 & 28.7 & 52.5\\
     & Sol & 26.3 & 16.2 & 36.3\\
     & Terra & 26.2 & 17.5 & 36.3\\
     & Fable~5.1 & 57.5 & 46.2 & 68.8\\
     & Grok~4.7 & 12.5 & 1.2 & 23.8\\
    \bottomrule

%% file: tables/tab_app_class_body.tex
    Claude Opus~5.5 & 100 & 25 & 100 & 85 && 100 & 96 & 100 & 100 && 92 & 36 & 85 & 100\\
    GPT-6 Astra & 91 & 75 & 100 & 90 && 100 & 100 & 100 & 100 && 90 & 33 & 54 & 77\\
    Claude Sonnet~5.5 & 98 & 18 & 100 & 70 && 100 & 25 & 100 & 100 && 90 & 26 & 15 & 73\\
    Claude Haiku~5.5 & 98 & 5 & 90 & 60 && 100 & 17 & 94 & 97 && 49 & 3 & 12 & 58\\
    GPT-6 Luna & 95 & 2 & 95 & 20 && 100 & 21 & 100 & 100 && 74 & 5 & 4 & 27\\
    GPT-6.1 Sol & 100 & 66 & 100 & 75 && 100 & 100 & 100 & 100 && 87 & 28 & 54 & 77\\
    GPT-6.1 Terra & 100 & 50 & 100 & 70 && 100 & 88 & 100 & 97 && 85 & 33 & 27 & 73\\
    Claude Fable~5.1 & 91 & 86 & 100 & 85 && 100 & 83 & 100 & 100 && 87 & 18 & 54 & 100\\
    Gemini~4 Argon & 95 & 74 & 100 & 94 && 100 & 91 & 100 & 94 && 92 & 31 & 42 & 23\\
    Gemini~4 Carbon & 93 & 66 & 95 & 90 && 100 & 29 & 100 & 78 && 100 & 0 & 96 & 15\\
    Grok~4.7 & 95 & 23 & 100 & 60 && 100 & 12 & 97 & 100 && 69 & 10 & 46 & 46\\
    Nova~2 Lite & 93 & 7 & 79 & 30 && 83 & 29 & 53 & 64 && 10 & 0 & 12 & 4\\
    Kimi K2.5$^\dagger$ & 91 & 0 & 90 & 30 && 100 & 17 & 75 & 78 && 26 & 5 & 19 & 15\\
    Qwen3-VL-235B$^\dagger$ & 98 & 2 & 100 & 25 && 100 & 8 & 44 & 44 && 22 & 3 & 12 & 15\\
    Gemma~3~27B$^\dagger$ & 98 & 0 & 65 & 5 && 92 & 25 & 42 & 28 && 33 & 8 & 4 & 8\\
    Ministral~3~14B$^\dagger$ & 72 & 5 & 85 & 30 && 58 & 25 & 64 & 42 && 10 & 3 & 19 & 8\\
    \bottomrule